\documentclass[conference]{IEEEtran}
\IEEEoverridecommandlockouts
\usepackage{cite}
\usepackage{amsmath,amssymb,amsfonts}
\usepackage{algorithmic}
\usepackage{graphicx}
\usepackage{textcomp}
\usepackage{xcolor}
\usepackage{svg}
\usepackage{underscore}
\usepackage{makecell}
\svgsetup{
    inkscapepath=i/svg-inkscape/
}
\svgpath{{svg/}}
\def\BibTeX{{\rm B\kern-.05em{\sc i\kern-.025em b}\kern-.08em
    T\kern-.1667em\lower.7ex\hbox{E}\kern-.125emX}}
\begin{document}

\title{An improved helmet detection method for YOLOv3 on an unbalanced dataset}

\author{
\IEEEauthorblockN{Rui Geng}
\IEEEauthorblockA{School of Software Technology \\
Dalian University of Technology\\
Dalian, China \\
gengruigr@qq.com}
\and
\IEEEauthorblockN{Yixuan Ma}
\IEEEauthorblockA{School of Economics \\
Beijing International Studies University\\
Beijing, China \\
yixuanma9@outlook.com}
\and
\IEEEauthorblockN{Wanhong Huang}
\IEEEauthorblockA{School of Software Technology \\
Dalian University of Technology\\
Dalian, China \\
mannho_dut@yahoo.co.jp}
}

\maketitle

\begin{abstract}
The YOLOv3 target detection algorithm is widely used in industry due to its high speed and high accuracy, but it has some limitations, such as the accuracy degradation of unbalanced datasets. The YOLOv3 target detection algorithm is based on a Gaussian fuzzy data augmentation approach to pre-process the data set and improve the YOLOv3 target detection algorithm. Through the efficient pre-processing, the confidence level of YOLOv3 is generally improved by 0.01-0.02 without changing the recognition speed of YOLOv3, and the processed images also perform better in image localization due to effective feature fusion, which is more in line with the requirement of recognition speed and accuracy in production.
\end{abstract}


\section{Introduction}
As an effective protective tool, the safety helmet has been applied to all kinds of production sites, but due to the reason that the supervision and audit are not in place, accidents caused by not wearing the safety helmet occur from time to time. Therefore, the realization of helmet wearing detection is of great significance for the protection of the life safety of the staff at the production site.

In helmet detection, the main use is the target detection algorithm\cite{b1,b2}. In recent years, the target detection algorithm has become an important research hotspot in the field of computer vision research. There are two types of target detection methods based on deep learning: one is R-CNN target detection algorithm based on candidate regions\cite{b3,b4,b5,b6}, which needs to generate candidate regions first, and then do classification and regression operations on the candidate regions; the other is YOLO\cite{b13,b14,b15,b16,b19} (You Only Look Once) and SSD algorithm\cite{b2}, which uses only one CNN network to directly predict the category and location of different targets. Compared with the R-CNN algorithm, YOLO can achieve real-time detection speed, but the accuracy is lower.

In order to improve the accuracy of YOLO, Redmon et al.\cite{b21} proposed YOLOv2\cite{b20} and YOLOv3\cite{b14,b21}, which improve the prediction accuracy while maintaining the speed advantage, especially for the identification of small objects. However, it is found that the accuracy of YOLOv3 decreases when the data set is unbalanced and irregular, and the work in this paper is to make improvements on the basis of YOLOv3 so that it can perform better on irregular and unbalanced data sets. The full source code of the project is available at https://github.com/ridin/YOLOv3-on-an-unbalanced-dataset.

\section{Relevant research}

\subsection{Region CNN and YOLO}

Silva et al.\cite{b11} used the Circle Hough Transform (CHT) and Histogram of Oriented Gradient (HOG) descriptors to extract image features, and a multilayer perceptron machine (Multi-layer Perceptron, MLP) to classify the target. This method works well for single-worn detection, but it is poorly effective for multi-worn detection and cannot be applied to multi-person images. In recent years, deep learning-based target detection techniques have developed rapidly, which are mainly divided into two types, namely the two-stage method based on region suggestion and the one-stage detection method without region suggestion.

The two-stage target detection algorithm is mainly RCNN (Regions with Convolutional Neural Network) columns, such as RCNN, Fast RCNN (Fast Regions with Convolutional Neural Network feature), and Faster-RCNN, with Faster-RCNN having the best detection performance.

One-stage target detection algorithm Among them, the SSD (Single Shot MultiBox Detector) and YOLO algorithms are representative, such as YOLO, YOLOv2, and YOLOv3, with YOLOv3 being the most effective. Based on YOLOv3, an image pyramid pattern is used to acquire feature maps at different scales, and multi-scale training is used to increase the adaptability of the model and improve the efficiency and accuracy of helmet wearing detection. Both traditional and deep learning-based methods have achieved relatively good results in helmet wearing detection, but there are problems of reduced accuracy and difficulty in improving the upper limit of accuracy for unbalanced datasets\cite{b8,b18,b12}.

\subsection{YOLOv3}

The YOLOv3 network uses Darknet53 as the feature extraction network, in YOLOv3, there is only a convolutional layer, most of them are 3*3 convolution, and compared to YOLOv2, the pooling layer is canceled, and the size of the output feature map can be controlled by adjusting the step size of the convolutional layer. YOLOv3 borrows the idea of a "pyramid feature map" (FPN) and uses small feature maps to detect large objects, while large feature maps are used to detect large objects. The reason for this approach is that the lower-level network features have more precise location characteristics, while the higher-level network features have richer semantic information.

The feature map used for prediction by YOLOv3 has both accurate location information of low-level network feature map and rich semantic information of high-level network feature map, which enables the prediction layer of YOLOv3 to both locate the object accurately and identify the category of the object correctly. The output dimension of the feature map is $N*N*[3*(4+1+M)]$, where $N*N$ is the number of output grid, and each grid has 3 Anchor boxes, each box has a 4-dimensional prediction value box $t_x$, $t_y$, $t_w$, $t_h$, and 1-dimensional object prediction box confidence level and M-dimensional object category number. In category prediction, YOLOv3 uses multiple logical classifiers instead of softmax to classify each box and uses a binary cross-entropy loss function to predict the categories during training.

\section{Our approach}
We first perform feature extraction to determine the distribution and mathematical characteristics of the dataset; then we build YOLOv3 on Mxnet for training on PC; then we use a test set of 689 images with a size of 95.4 MB provided by the competition, and finally, we export the test results to a txt file.

\subsection{Feature extraction statistics (unbalanced dataset)}\label{AA}
Based on the features of the dataset, we can obtain relevant information that will provide better support in building neural network training. For feature extraction, the steps are as follows.
\begin{itemize}
\item Calculate the proportion of each target in the original image.
\item Calculate the average length of the target.
\item Calculate the average width.
\item Calculate the average area.
\item Calculate the average percentage of the target.
\end{itemize}

After the above process, we get the following description: the dataset consists mainly of people wearing helmets and a wide variety of backgrounds. The size of the dataset is 1.1 GB and there are 7581 images, examples of which are as follows.

\begin{figure}[htbp]
\centerline{\includegraphics[height=4.5cm,width=7.5cm]{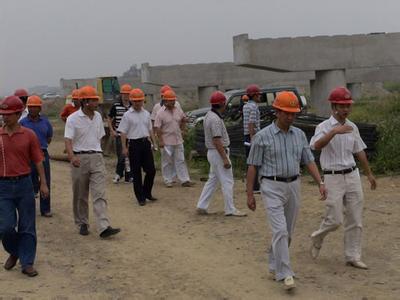}}
\caption{Image in the dataset}
\label{fig}
\end{figure}

After processing, we found that there is an imbalance in the dataset. An unbalanced dataset refers to a dataset with extremely unbalanced sample sizes for each category. Take the binary classification problem as an example, e.g., the number of samples in positive categories is much larger than the number of samples in negative categories. The treatment of unbalanced data sets is divided into two aspects:

\begin{itemize}

\item From the perspective of data, the main method is sampling, which is divided into under-sampling and oversampling and some corresponding improvement methods.

\item From the perspective of algorithms, the algorithms are optimized by considering the differences in the cost of different misclassification situations. In the field of image recognition, we use image enhancement processing methods to process the relevant images.

\end{itemize}

\begin{figure}[htbp]
\centerline{\includegraphics[height=4.5cm,width=7.5cm]{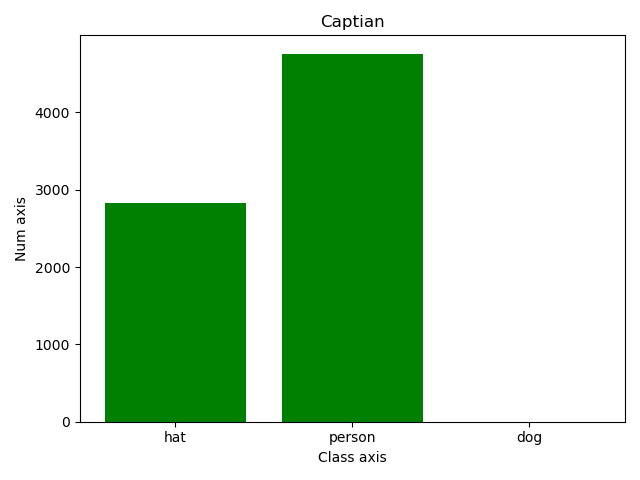}}
\caption{Ratio of hat to person}
\label{fig}
\end{figure}
\subsection{Pre-processing and training}
In the training, we processed the process as shown in the figure below, and now we will explain our processing in steps.
\begin{itemize}
\item MXNet gets the data and reads it into the VOC dataset.
\item Test on validation dataset, set nms threshold and topk constraint.
\item Training pipeline: Download the data, decompress the data, and divide the data into different directories for training according to tags.
\item Cross entropy as a loss function.
\item Set target matrix and center matrix.
\item Circuit training.
\end{itemize}

\subsection{Read the data set using the method provided by MXNet for reading VOC datasets}
Many deep learning platforms have computer vision APIs that provide common vision models and their pre-training weights, which greatly facilitate our learning and experimentation. . We used the GluonCV framework.
For GluonCV, we just need to prepare the general format (.rec) dataset and feel free to experiment with the visual model in model_zoo. So a first task is to convert our own dataset to the MXNet preferred .rec binary.

The result after reading the common image is to form a directory structure is to separate the image and txt form of the label file into two folders, named images, and labels respectively. in the root directory to generate a RecDataSet folder, which stores yolo.lst, yolo.idx, yolo.rec three files; then converted to VOC Our directory structure is to put all the images and labeled xml files with the same name in the same folder, then run the data read-in code of the project; finally, generate a folder of RecDataSet in the root directory, where the voc.lst, voc.idx, and voc.rec files are stored.

\subsection{Image processing}
For the unbalanced dataset, we use Gaussian blurring means for image enhancement. The process is rough as follows: build an image loader and apply image transformers to the loader for image enhancement. This is the way to deal with the data imbalance problem.
Gaussian blur is a processing technique used to eliminate noise, reduce the level of detail, and blur the image. Gaussian blur is also a template for the weighted average method and uses a positive distribution as a template for pixel mapping work. The formula is as follows:

N dimensional space:

\begin{equation}
G(r) = \frac{1}{\sqrt{2\pi\sigma^2}^N}e^{-r^2/(2\sigma^2)}
\end{equation}

% \begin{figure}[htbp]
% \centerline{\includegraphics[height=1.6875cm,width=2.8125cm]{2.png}}
% \end{figure}

Two-dimensional space:
\begin{equation}
G(u,v) = \frac{1}{2\pi\sigma^2}e^{-(u^2+v^2)/(2\sigma^2)}
\end{equation}
%
% \begin{figure}[htbp]
% \centerline{\includegraphics[height=1.6875cm,width=2.8125cm]{3.png}}
% \end{figure}
Since the closer the image is to the object being processed, the greater its effect on that pixel, we use a weighted average when Gaussian blurring. Blurring can be understood as each pixel taking the average value of the surrounding pixels. The middle point takes the average of the surrounding points, and it becomes the average. Numerically, this is a smoothing. On a graphic, it is equivalent to creating a blurring effect, where the midpoints lose detail. In actual image processing, a nine-pixel grid is formed around each center point. And the center position is taken as (0,0), then the other positions are shown as follows:

\begin{table}[htbp]
\caption{Position}
\begin{center}
\setlength{\tabcolsep}{11mm}
\begin{tabular}{|c|c|c|}
\hline
\textbf{\textit{(-1,1)}}& \textbf{\textit{(0,1)}}& \textbf{\textit{(1,1)}} \\
\cline{1-3}
\textbf{\textit{(-1,0)}}& \textbf{\textit{(0,0)}}& \textbf{\textit{(1,0)}} \\
\hline
\textbf{\textit{(-1,-1)}}& \textbf{\textit{(0,-1)}}& \textbf{\textit{(1,-1)}}  \\
\hline
\end{tabular}
\label{tab1}
\end{center}
\end{table}

Based on this position information, calculate the weight matrix needs to set the weight value $\alpha$ by itself, get the weight matrix, calculate the weighted average of these 9 points, let their weight sum is equal to 1. For the actual image of the grayscale value, the image as each point multiplied by its weight value, add these 9 values, is the value of the Gaussian blur of the central point. Repeat this process for all points to get the Gaussian blurred image. If the original image is a color image, you can do a Gaussian blur for each of the three RGB channels. Set the grayscale value of each grid to $a_n$ and the weight value to $r_n$(n=1,2,...9), then the actual processed effect is as follows:

\begin{table}[htbp]
\caption{Gaussian value}
\begin{center}
\setlength{\tabcolsep}{11mm}
\begin{tabular}{|c|c|c|}
\hline
\textbf{\textit{a$_1$*r$_1$}}& \textbf{\textit{a$_2$*r$_2$}}& \textbf{\textit{a$_3$*r$_3$}} \\
\cline{1-3}
\textbf{\textit{a$_4$*r$_4$}}& \textbf{\textit{a$_5$*r$_5$}}& \textbf{\textit{a$_6$*r$_6$}} \\
\hline
\textbf{\textit{a$_7$*r$_7$}}& \textbf{\textit{a$_8$*r$_8$}}& \textbf{\textit{a$_9$*r$_9$}}  \\
\hline
\end{tabular}
\label{tab1}
\end{center}
\end{table}

Then the Gaussian value of the center point is 9 values summed, which in turn fills the image with new grayscale values to get the blurred image. Using the blurred image, we train to better circumvent the effects of imbalance in the dataset.

\subsection{Testing on the validation dataset, setting nms threshold and topk constraint}

This step is to ensure the consistency and accuracy of the subsequent training images. Set nms_thresh=0.45, nms_topk=400 to validate; after validation, crop the image size, then complete the operation of split ground truths, and finally use eval_metric.update() to evaluate the cut image and update parameters.

\subsection{Training pipeline: the data is divided into different directories according to labels for training.}

Loading MXNet's Yolo model for training, with lr set to 0.001, epoch set to 10, the optimization function selected SGD, and the activation function selected sigmoid.
Loading of the MXNet model for predictions, with the predictions stored in txt (format: "Label Confidence Left Top Right Button").
Use arg parse to provide command-line parameter settings, as follows.

\begin{itemize}
\item Create the ArgumentParser() object.
\item Call the add_argument() method to add an argument.
\item Use parse_args() to parse the added parameters.
\end{itemize}

The constitutive neural network is shown

\begin{figure}[htbp]
\centerline{\includegraphics[height=5.5cm,width=8.7cm]{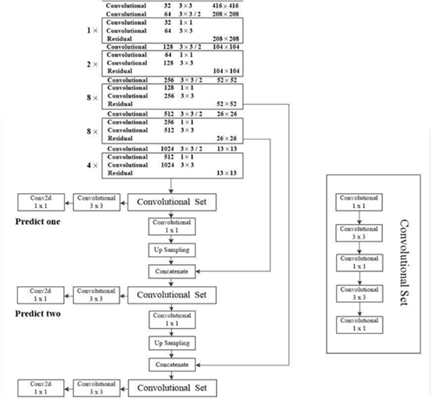}}
\caption{Neural networks}
\label{fig}
\end{figure}

Using cross-entropy as a category classification, define the SigmoidBinaryCrossEntropyLoss() method, using the following equation

max(x, 0) - x * z + log(1 + exp(-abs(x))))

\subsection{Loss function definition}

In YOLOv3, Loss is divided into three sections.

\begin{itemize}
\item The error introduced by the xywh part, i.e., the loss introduced by the b$_box$, is calculated using the following mathematical method.

% \begin{figure}[htbp]
% \centerline{\includegraphics[height=1.6875cm,width=2.8125cm]{5.png}}
% \end{figure}
\begin{equation}
  \mathit{IoU = \frac{|A\cap B|}{|A\cup B|}}
\end{equation}

Using Glou to calculate l$_box$:

% \begin{figure}[htbp]
% \centerline{\includegraphics[height=1.6875cm,width=2.8125cm]{6.png}}
% \end{figure}
\begin{equation}
  \mathit{GIoU = IoU - \frac{|A_c-U|}{|A_c|}}
\end{equation}

where A$_c$ represents the area of the smallest closed area of the two boxes, that is, the area of the smallest box that contains both the predicted and real boxes.

\item The error brought by the confidence level, i.e., the loss brought by obj is divided into Center and Scale. l$_obj$ represents the probability of whether the bounding box contains objects or not. In the implementation, obj loss can be specified by arc, and we use default mode: using BCEWithLogitsLoss, obj loss is calculated separately from cls loss.

\item The error caused by the class, that is, the loss caused by the class. here called l$_cls$, using default mode, using BCEWithLogitsLoss to calculate the class loss.

\end{itemize}

The final loop training gets the model.

\section*{Test results}

To detect the improvement of the Gaussian fuzzy optimization method on the recognition accuracy of the unbalanced dataset, we introduce the comparison of the recognition accuracy of the same image before and after the Gaussian fuzzy optimization. The results are presented as a text document for each image, with one detection per column in the document, e.g., "Label Confidence, Left, Top, Right, Bottom". If a picture has more than one detection, the use of more than one line out. Here is a picture as an example, in the local PC test, to show the test results, use the following picture.

\begin{figure}[htbp]
\centerline{\includegraphics[height=4.5cm,width=7.5cm]{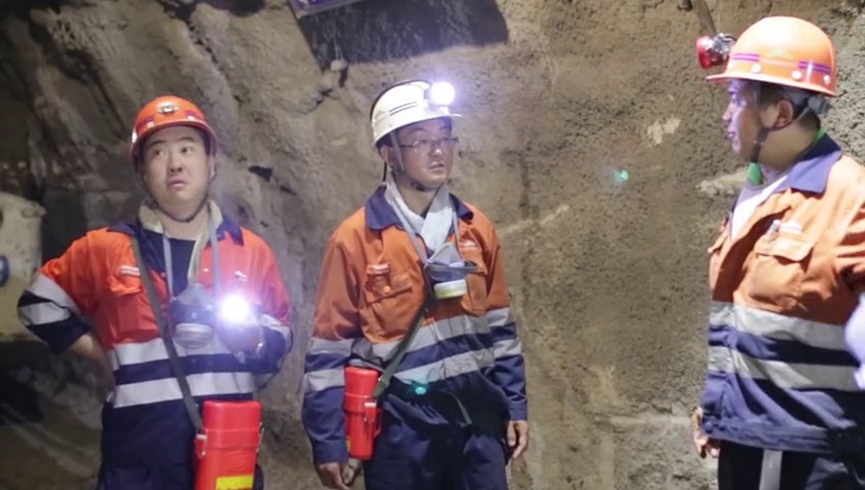}}
\caption{Image in the test set}
\label{fig}
\end{figure}

The following results were obtained.

\begin{table}[htbp]
\caption{YOLOv3}
\begin{center}
\begin{tabular}{|c|c|c|c|c|c|}
\hline
\multicolumn{6}{|c|}{\textbf{YOLOv3}} \\
\cline{1-6}
\textbf{\textit{category}}& \textbf{\textit{Label Confidence}}& \textbf{\textit{Left}}& \textbf{\textit{Top}}& \textbf{\textit{Right}}& \textbf{\textit{Bottom}} \\
\cline{1-6}
\textbf{\textit{hat}}& \textbf{\textit{0.981}}& \textbf{\textit{598.40}}& \textbf{\textit{5.74}}& \textbf{\textit{718.82}}& \textbf{\textit{143.52}} \\
\cline{1-6}
\textbf{\textit{hat}}& \textbf{\textit{0.965}}& \textbf{\textit{106.98}}& \textbf{\textit{74.34}}& \textbf{\textit{188.41}}& \textbf{\textit{177.79}} \\
\hline
\textbf{\textit{hat}}& \textbf{\textit{0.968}}& \textbf{\textit{314.11}}& \textbf{\textit{60.32}}& \textbf{\textit{395.96}}& \textbf{\textit{158.78}} \\
\hline
\end{tabular}
\label{tab1}
\end{center}
\end{table}

\begin{table}[htbp]
\caption{Improved YOLOv3}
\begin{center}
\begin{tabular}{|c|c|c|c|c|c|}
\hline
\multicolumn{6}{|c|}{\textbf{Improved YOLOv3}} \\
\cline{1-6}
\textbf{\textit{category}}& \textbf{\textit{Label Confidence}}& \textbf{\textit{Left}}& \textbf{\textit{Top}}& \textbf{\textit{Right}}& \textbf{\textit{Bottom}} \\
\cline{1-6}
\textbf{\textit{hat}}& \textbf{\textit{0.999}}& \textbf{\textit{598.08}}& \textbf{\textit{5.79}}& \textbf{\textit{718.31}}& \textbf{\textit{143.43}} \\
\cline{1-6}
\textbf{\textit{hat}}& \textbf{\textit{0.995}}& \textbf{\textit{106.91}}& \textbf{\textit{74.47}}& \textbf{\textit{188.49}}& \textbf{\textit{177.79}} \\
\hline
\textbf{\textit{hat}}& \textbf{\textit{0.977}}& \textbf{\textit{314.10}}& \textbf{\textit{60.35}}& \textbf{\textit{395.93}}& \textbf{\textit{158.73}} \\
\hline
\end{tabular}
\label{tab1}
\end{center}
\end{table}

To validate the effectiveness of the improved method proposed in this paper, it is further validated on 689 test data. This section realizes the whole test results on YOLOv3 and improved YOLOv3 at the same time. The detection results of different improved models on the experimental data are shown in the table.

For the 689 test set data, we used YOLOv3 and the improved YOLOv3 for the helmet wearing detection. It can be observed that the improved YOLOv3 can detect helmet wearers well compared to YOLOv3, and can detect more targets with accurate locations and complete bounding boxes.

\begin{table}[htbp]
\caption{Overall test results for both models}
\begin{center}
\setlength{\tabcolsep}{11mm}
\begin{tabular}{|c|c|}
\hline
\cline{1-2}
\textbf{\textit{Model}}& \textbf{\textit{Label Confidence}} \\
\cline{1-2}
\textbf{\textit{YOLOv3}}& \textbf{\textit{0.966}} \\
\cline{1-2}
\textbf{\textit{Improved YOLOv3}}& \textbf{\textit{0.982}}\\
\hline
\end{tabular}
\label{tab1}
\end{center}
\end{table}

It can be seen that the confidence level, which is used as a metric to evaluate the accuracy, has a general improvement of 0.01-0.02 for helmet identification after the Gaussian fuzzy optimization mentioned in this paper, which is important to improve the ultimate accuracy of YOLOv3 in this state; at the same time, after the Gaussian fuzzy processing, due to the effective feature fusion, the low-level feature map contains semantic information and the high-level feature map contains more detailed information, which improves the recall rate of small-sized helmets and the accuracy of large-sized helmet localization.

\section*{Related work}

For Gaussian blur, Jan Flusser et al.\cite{b9} introduced the notion of a native image as the normative form of all Gaussian blur-equivalent images. The native image is defined in the spectral domain by projection operators, and they proved that the moments of the native image are the normative form of all Gaussian blur-equivalent images. They prove that the moments of the image is invariant to Gaussian blur and derive recursive formulas for the direct computation of the image, without actually constructing the image itself\cite{b9}. In this paper, we draw on the idea of image and use recursion in Gaussian blurring to ensure that the pixels of the original image are invariant to the information after the operation of the Gaussian matrix.

Meanwhile, Niranjan D. Narvekar et al.\cite{b10} propose a reference-free image blurring metric based on human studies of the fuzzy perception of different contrast values. The metric uses a probabilistic model to estimate the probability of detecting blur for each edge in an image and then pools the information by calculating the cumulative probability of blur detection (CPBD)\cite{b10}. This approach inspired us to adopt an image fuzziness metric to deal with imbalances in the dataset. We introduce a fuzzy metric for processing raw images, an approach that ensures that the neural network's perception of the image is not lost when balancing the classification problem and still retains enough information that can be used for classification.

For YOLOv3, Fan Wu et al.\cite{b15,b22} propose a YOLOv3-based full regression deep neural network architecture that takes advantage of Densenet's advantages in model parameters and technical cost to replace the backbone of the YOLOv3 network for feature extraction\cite{b7,b14,b17,b21,b22}, resulting in the so-called YOLO-Densebackbone convolutional neural network. In this paper, we draw on the ideas of Fan Wu et al.\cite{b15,b22} to optimize the construction of the convolutional layer, the cross-entropy, and the selection of the loss function.

\section*{Improvement Methods}

For the helmet detection problem, there are several techniques we can use to continue to improve accuracy. For example, for problems where there is not enough light in the production situation and the target occupies a small frame, we can improve algorithms such as YOLOv3 or Faster RCNN for processing such as feature fusion, multiscale detection, and increasing the number of anchor points. At the same time, we can also introduce more deep learning techniques in algorithms such as YOLOv3 or Faster RCNN, such as adding multilayer convolutional feature fusion techniques in building neural networks, and introducing online difficult sample mining methods in target detection to enhance the robustness of small and occluded targets in different environments, which are the directions of our future in-depth research.

\section*{Conclusion}

The direct use of YOLOv3 can no longer adapt to the real production situation for target detection requirements. For the unbalanced helmet dataset, we use data augmentation to process the data and then train the model with YOLOv3 algorithm, without affecting the recognition speed, and finally get a 0.01-0.02 increase in the confidence level. The image detection and classification problems similar to helmet detection can draw on the results in this thesis to complete the further optimization of the recognition effect, which is of great significance for the rapid image detection and recognition of actual production sites.

\vspace{12pt}
\end{document}